\documentclass[10pt,twocolumn,letterpaper]{article}

\usepackage[pagenumbers]{iccv} 

\definecolor{iccvblue}{rgb}{0.21,0.49,0.74}
\usepackage[pagebackref,breaklinks,colorlinks,allcolors=iccvblue]{hyperref}

\title{Efficient Multi-Camera Tokenization with Triplanes for End-to-End Driving}

\author{Boris Ivanovic$^1$ \hspace{0.25cm} Cristiano Saltori$^1$ \hspace{0.25cm} Yurong You$^1$ \hspace{0.25cm} Yan Wang$^1$ \hspace{0.25cm} Wenjie Luo$^1$ \hspace{0.25cm} Marco Pavone$^{1,2}$\\
$^1$NVIDIA Research \hspace{0.5cm} $^2$Stanford University\\
{\tt\small \{bivanovic,csaltori,yurongy,yanwan,wenjiel,mpavone\}@nvidia.com, pavone@stanford.edu}
}

\begin{document}
\maketitle
\begin{abstract}

Autoregressive Transformers are increasingly being deployed as end-to-end robot and autonomous vehicle (AV) policy architectures, owing to their scalability and potential to leverage internet-scale pretraining for generalization. Accordingly, tokenizing sensor data \emph{efficiently} is paramount to ensuring the real-time feasibility of such architectures on embedded hardware. To this end, we present an efficient triplane-based multi-camera tokenization strategy that leverages recent advances in 3D neural reconstruction and rendering to produce sensor tokens that are agnostic to the number of input cameras and their resolution, while explicitly accounting for their geometry around an AV. Experiments on a large-scale AV dataset and state-of-the-art neural simulator demonstrate that our approach yields significant savings over current image patch-based tokenization strategies, producing up to 72\% fewer tokens, resulting in up to 50\% faster policy inference while achieving the same open-loop motion planning accuracy and improved offroad rates in closed-loop driving simulations.

\end{abstract}    
\section{Introduction}
\label{sec:intro}

Generalizing to novel environments remains a grand challenge for robots and autonomous vehicles (AVs), requiring advanced reasoning capabilities to robustly handle unseen scenarios.
Accordingly, coinciding with the rapid advancement of
token-based autoregressive (AR) Transformers~\cite{VaswaniShazeerEtAl2017} such as LLMs and VLMs,
there has been a significant growth in research aiming to endow robots and AVs with general reasoning capabilities by deploying internet-pretrained foundation models on-vehicle.
However, such architectures oft contain billions of parameters, complicating their real-time deployment in embedded systems~\cite{gao2024avfmsurvey}.

Many approaches are currently being investigated to reduce model size and inference time, including model compression~\cite{zhu2024compression} and efficient sensor tokenization strategies.
In particular, image tokenizers aim to represent images with as few latent embeddings (i.e., ``tokens") as possible. Current methods largely focus on representing single images
and employ autoencoding architectures~\cite{SohnLeeEtAl2015,OordVinyalsEtAl2018,esser2021vqgan} or directly encode patches of pixels~\cite{dosovitskiy2020image}.
VLMs primarily employ Vision Transformers (ViTs)~\cite{dosovitskiy2020image} and adopt the latter, partitioning images into patches that are encoded to form a 1D token sequence. While this approach is practical, it produces token counts that scale linearly with image resolution and the number of cameras~\cite{wang2025patchscaling}.
To obtain a 360-degree view of their surroundings, AVs often use 6 to 10 cameras, the patch-based tokenization of which would
yield thousands of tokens per timestep, precluding real-time inference.

\begin{figure}[t]
  \centering
  \includegraphics[width=\linewidth]{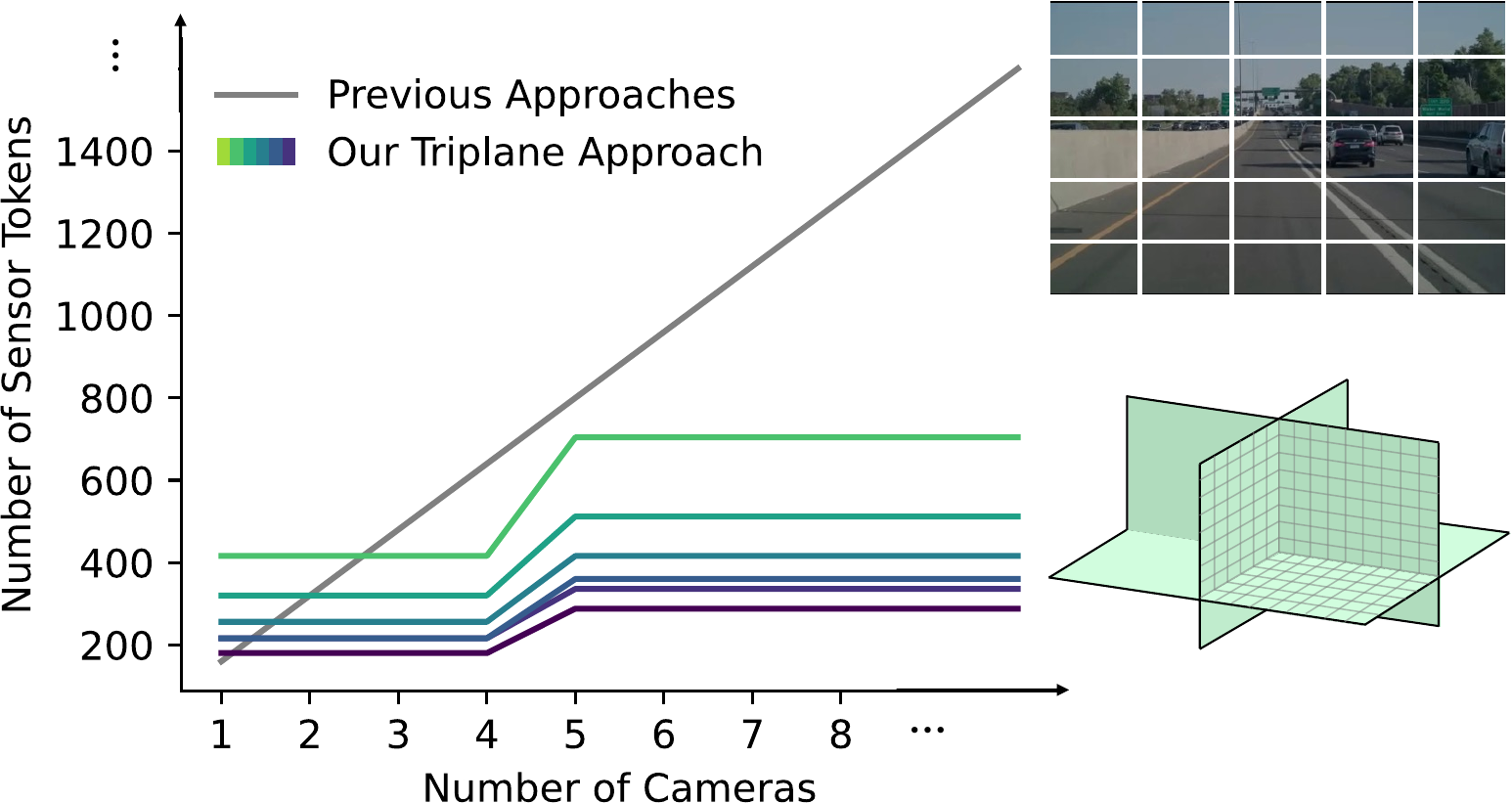}
   \caption{Many Transformer-based AV policies employ patch-based image tokenization, whose token counts scale linearly with the number of cameras, complicating real-world AV deployment. In contrast, our proposed triplane-based approach produces a \emph{fixed} number of tokens, irrespective of the number of cameras or their resolution, significantly reducing online inference time.}
   \label{fig:hero}
\end{figure}

\textbf{Contributions.} To address this, we present a novel multi-camera image tokenization scheme (\cref{fig:hero}) based on triplanes, a recent advancement in 3D neural reconstruction and rendering, that produces tokens in a resolution-agnostic, camera-number-agnostic, and geometrically-aware manner (\cref{sec:method}).
Experiments on a large-scale dataset consisting of 20,000 hours of driving data from
25 countries demonstrate the efficacy of our approach over patch-based tokenization, producing up to 72\% fewer tokens, resulting in up to 50\% faster inference while achieving the same open-loop motion planning accuracy and improved offroad rates in closed-loop driving simulations (\cref{sec:expts}).

\section{Related Work}
\label{sec:related_work}

\textbf{End-to-End Driving with Tokens.} The first approaches applying token-based Transformer~\cite{VaswaniShazeerEtAl2017} architectures to end-to-end (E2E) driving primarily focused on connecting previously-existing modules (e.g., detection, prediction, planning) along a differentiable pathway, treating their outputs as queries, keys, and values into subsequent modules. Initial works UniAD~\cite{hu2023planning} and VAD~\cite{jiang2023vad} encode multi-camera observations into a birds-eye view (BEV) feature grid, where each cell is treated as a token and processed through a cascade of task-specific Transformer decoders (e.g., for object tracking, map segmentation, and motion prediction). PARA-Drive~\cite{weng2024paradrive} further tokenizes BEV feature grid cells into parallel multi-task decoding heads through a shared Transformer backbone. Each of these works are trained from scratch on AV-specific data and all model outputs are supervised with ground-truth (GT) labels.

Inspired by progress in language modeling, a parallel line of work aims to develop generalist embodied agents that directly use an AR Transformer~\cite{VaswaniShazeerEtAl2017} to process input text, actions, and sensor data into downstream task-specific tokens.
RT-1~\cite{brohan2022rt1}, Gato~\cite{reed2022gato}, RT-2~\cite{brohan2023rt2}, RT-X~\cite{openx2023rtx}, and VIMA~\cite{jiang2023vima} broadly introduce, extend, and deploy vision-language-action models (VLAMs) onto multiple robotic tasks and embodiments. 
Initial works in the AV domain focused on AR video generation~\cite{hu2023gaia1} and predicting text explanations alongside driving actions~\cite{wayveLINGO1,xu2024drivegpt4,wayveLINGO2} from a front-facing camera, tokenizing images with the patch-based embedding scheme from ViT~\cite{dosovitskiy2020image}.

Multiple cameras have recently been incorporated in VLM-based E2E AV models such as DriveVLM~\cite{tian2024drivevlm}, OmniDrive~\citep{wang2024omnidrive}, and EMMA~\cite{hwang2024emma}. DriveVLM adopts the ViT-based SigLIP~\cite{zhai2023siglip} to tokenize images, OmniDrive develops a sparse 3D query architecture to encode multiple cameras, and EMMA feeds camera images into Gemini Nano-1~\cite{gemini}. However, these approaches require a separate token reduction model (e.g., LDPNetv2~\cite{chu2023mobilevlm}) to reduce inference times, auxiliary training signals that are costly to obtain, or produce a large number of tokens (500+) per image. In contrast, our approach is fully self-supervised (it can optionally use auxiliary signals if provided) and produces much fewer tokens (\cref{sec:expts}).

\textbf{Latent Image Representations.} While current approaches mostly employ ViTs~\cite{dosovitskiy2020image}, another common tokenization strategy is to use the latent space of a pretrained image autoencoder. VAE~\cite{SohnLeeEtAl2015}, VQ-VAE~\cite{OordVinyalsEtAl2018}, and VQ-GAN~\cite{esser2021vqgan} have demonstrated that encoder-decoder architectures can successfully compresses high-dimensional images into a low-dimensional (optionally discrete) latent space, which can be used as tokens for E2E driving models. MoST~\cite{mu2024most} adopts this approach, tokenizing each image into 256 tokens using a VQ-GAN~\cite{esser2021vqgan} encoder (alongside other ViT-based models). TiTok~\cite{yu2024titok} combines ViTs and VQ-GANs into a unified framework that further pares down the number of tokens needed to as low as 32. 
More broadly, while autoencoders can represent images parsimoniously, they require a fixed input resolution and focus on single images (neglecting camera geometries and scaling linearly with camera count and resolution).

More recent works explore alternative image representations (e.g., tokenizing the frequency spectrum~\cite{esteves2024spectral}) and adaptive methods that allocate varying token amounts to images~\cite{duggal2025how,yan2025elastictok}. Such advancements are exciting, and orthogonal to our approach since it can be paired with any backbone image encoder or online token reducer.

\textbf{Volumetric Latent Representations.} Accounting for multi-camera geometry is a critical aspect of the AV domain, due to both the number of cameras and their orientation (pointing outwards). To this end, there have been many volumetric latent representations proposed for the neural reconstruction and rendering of 3D environments that can also be used for tokenization.

Explicit representations like voxels and 3D Gaussian Splatting~\cite{kerbl2023gs} can be produced in a feedforward, differentiable manner from input images, but they produce too many elements (ie., voxels and Gaussians) for direct tokenization. These can be alleviated with sparse voxel hierarchies~\cite{wang2024distillnerf} or Gaussian merging and pruning strategies, but they still result in thousands of tokens per scene. On the other hand, fully-implicit approaches such as NeRF~\cite{mildenhall2020nerf} and SDF~\cite{Park_2019_CVPR,takikawa2021nglod}
require little memory, but lack the ability to be produced from input images in a feedforward manner.

Hybrid explicit-implicit approaches such as $K$-planes~\cite{kplanes_2023,Cao2023HEXPLANE} lie in the middle, factorizing time and space into $K$ feature grids, with $K=3$ (triplanes) being the most common choice~\cite{Chan2022}.
Their main advantage is efficiency; the bulk of modeling capacity is shifted to explicit features, allowing for a smaller feature decoder at rendering time without losing expressiveness~\cite{Chan2022}.
Accordingly, triplanes have been widely used to represent faces~\cite{Chan2022,trevithick2023live3d,trevithick2024wysiwyg}, objects~\cite{Shue2023,hong2024lrm}, indoor environments~\cite{yan2024frankenstein}, and, more recently, urban driving scenes~\cite{huang2023tri,huang2024self,xu2025survey}.
While these works mainly use triplanes for neural rendering or occupancy modeling, we focus on their tokenization for use by an AR Transformer~\cite{VaswaniShazeerEtAl2017}.

\begin{figure*}
  \centering
  \includegraphics[width=\linewidth]{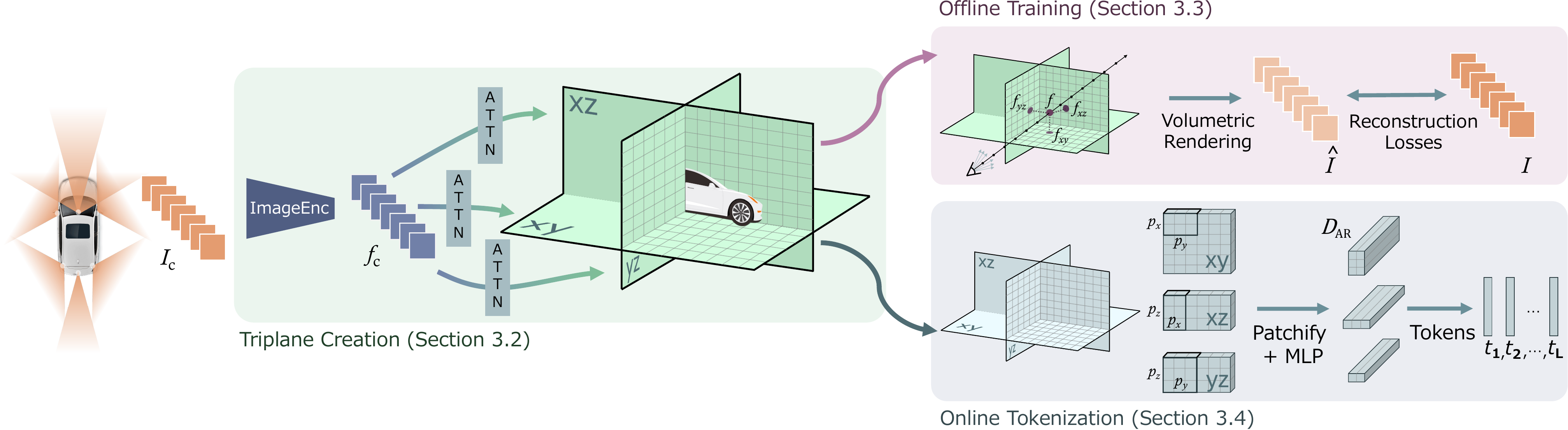}
  \caption{Our triplane-based multi-camera tokenization strategy. Images from multiple cameras are first encoded through ImageEnc, producing features which form triplanes through a set of geometric per-image and cross-image attention operations. The model is trained in a self-supervised manner, and yields triplanes that can be effectively tokenized for downstream use by an autoregressive Transformer.}
  \label{fig:method}
  \vspace{-0.4cm}
\end{figure*}
\section{Multi-Camera Tokenization with Triplanes}
\label{sec:method}

\subsection{Background: Triplanes}

Triplanes are a volumetric latent representation comprised of three axis-aligned orthogonal feature planes $P_{xy}, P_{xz}, P_{yz} \in \mathbb{R}^{S_{i} \times S_{j} \times D_f}$ (\cref{fig:method}), where $S_{i} \times S_{j}$ denotes each plane's spatial dimensions, with each grid cell corresponding to a $s_{i} \times s_{j}\ \text{m}^2$ region of space, and $D_f$ is the feature dimension. 
In this work, triplanes serve as a resolution-agnostic, camera-count-agnostic, and geometry-aware representation of multi-camera images, enabling efficient tokenization for use by AR Transformers. Our approach for doing so is illustrated in \cref{fig:method}.

\subsection{Encoding Multi-Camera Images into Triplanes}
\label{sec:method_encoding}

At time $t$, we assume an AV obtains an image $I_c \in \mathbb{R}^{H \times W \times D_\text{Ch}}$ from each of its $N$ cameras $c \in \{C_1, \dots, C_N\}$. To encode these $N$ images into a triplane $\{P_{xy}, P_{xz}, P_{yz}\}$, we first featurize images with a backbone encoder network 
\begin{equation}
\label{eqn:featurize}
f_c = \text{ImageEnc}(I_c)\ \forall\ c \in \{C_1, \dots, C_N\},
\end{equation}
where $f_c \in \mathbb{R}^{H_f \times W_f \times D_f}$ and $H_f \times W_f$ denotes the features' spatial size. Then, similar to TPVFormer~\cite{huang2023tri}, a grid of 3D query points $q \in \mathbb{R}^{S_x \times S_y \times S_z \times D_f}$ (where $S_x, S_y, S_z$ are the number of triplane grid cells per spatial dimension) attend to the image features through a series of per-image and cross-image deformable attention operations, leveraging each camera's intrinsic and extrinsic parameters for 3D-to-2D projections. Internally, a sinusoidal positional encoding~\cite{VaswaniShazeerEtAl2017} is used to represent triplane grid locations. Finally, the updated queries are averaged along each spatial dimension to produce the desired triplane $\{P_{xy}, P_{xz}, P_{yz}\}$.

Crucially, since the triplane sizes $S_x, S_y, S_z$ are fixed, this process decouples the number of cameras $N$ and their resolution $H \times W$ from the resulting number of tokens. 

Finally, as driving scenes are unbounded, we employ a nonlinear scene parametrization as in Mip-NeRF 360~\cite{barron2022mipnerf360} to model far-away objects without increasing triplane sizes. Specifically, using the $x$ axis as an example,
a grid coordinate $p_{g,x} \in \mathbb{R}$ is mapped to ego-relative coordinates $p_{\text{ego},x} \in \mathbb{R}$ with a symmetric bilinear grid resolution:
\begin{equation*}
p_{\text{ego},x} = \begin{cases}
    R_{o,x} (p_{g,x} + S_{\text{in},x}) - R_{i,x} S_{\text{in},x}, \ \text{if}\ p_{g,x} < -S_{\text{in},x}\\
    R_{i,x} p_{g,x}, \hspace{1.90cm} \text{if}\ -S_{\text{in},x} \leq p_{g,x} \leq S_{\text{in},x}\\
    R_{o,x} (p_{g,x} - S_{\text{in},x}) + R_{i,x} S_{\text{in},x}, \hspace{0.35cm} \text{if}\ p_{g,x} > S_{\text{in},x}
\end{cases}
\end{equation*}
where $R_{i,x}, R_{o,x} \in \mathbb{R}$ denote the resolutions (m / cell) of the inner and outer cells of the triplane, respectively, and $S_{\text{in},x} \in \mathbb{N}$ is the number of inner grid cells.

\subsection{Scalable Training via Volumetric Rendering}
\label{sec:method_training}

In contrast to prior works which leverage multiple complex (e.g., Hessian-based~\cite{huang2024self}) self-supervised losses, our work is trained to minimize only two pixel-wise reconstruction losses. Specifically,
\begin{equation}
\label{eqn:loss}
    \mathcal{L}(I, \hat{I}) = \lambda_\text{LPIPS} \text{LPIPS}(I, \hat{I}) + \lambda_1 \|I - \hat{I}\|_1
\end{equation}
where LPIPS is the learned perceptual image patch similarity metric~\cite{zhang2018perceptual}, $\lambda$ are scalar weighting factors, $\|\cdot\|_1$ is the $\ell_1$ norm, $\hat{I}$ is the predicted (rendered) image from the triplane, and $I$ denotes the ground truth (GT) image from data.

Images $\hat{I}$ are rendered from triplanes via ray sampling and aggregation. Triplanes are queried for 3D positions $x \in \mathbb{R}^3$ by first projecting $x$ onto each of the three feature planes, retrieving the corresponding feature vectors $f_{xy}, f_{xz}, f_{yz}$ via bilinear interpolation, and aggregating the three feature vectors into $f$ via elementwise product. A lightweight MLP then decodes the 3D features $f$ into color and density, which are rendered into RGB images using volumetric rendering~\cite{mildenhall2020nerf}.

Notably, in contrast to current autoencoder-based methods~\cite{esser2021vqgan,chang2022maskgit,yu2022vectorquantized}, our approach does not use any GAN losses (whose training process is particularly sensitive). As we will show in \cref{sec:expts}, even without these losses, our triplane-based driving model matches/exceeds the driving performance of autoencoder-based approaches.

\subsection{Tokenizing Triplanes for Downstream Use}
\label{sec:method_tokenizing}

With a trained model in hand, multi-camera input images can be converted into triplanes in a feedforward manner. Once a triplane is obtained, it can then be tokenized and ingested by a downstream AR Transformer.

\textbf{Continuous Tokens.} Triplanes can be converted into a 1D token sequence in multiple ways. In this work, we patchify them as in ViT~\cite{dosovitskiy2020image}, splitting triplanes into patches of features and encoding them into a set of feature vectors with the desired number of dimensions for downstream use.

Formally, each feature plane $P_{ij} \in \mathbb{R}^{S_{i} \times S_{j} \times D_f}$ is converted to a token sequence $\{t_\ell\}$ by first reshaping it into a set of patches $P_{ij}' \in \mathbb{R}^{S_i / p_i \times S_j / p_j \times D_f p_i p_j}$. Then, a single-layer MLP converts the resulting feature dimension $D_f p_i p_j$ into $D_\text{AR}$, producing a set of tokens $\{t_\ell\} \in \mathbb{R}^{L_{ij} \times D_\text{AR}}$ where $L_{ij} = \frac{S_i S_j}{p_i p_j}$ is the number of tokens needed to represent plane $P_{ij}$ and $D_\text{AR}$ is the feature dimension of the downstream AR Transformer. Performing this for each plane results in an overall token sequence $\{t_\ell\} \in \mathbb{R}^{L \times D_\text{AR}}$ where $L = L_{xy} + L_{xz} + L_{yz}$.

\textbf{Discrete Tokens.} While the above produces continuous tokens, discrete tokens can also be output. In particular, a Finite Scalar Quantization (FSQ)~\cite{mentzer2024fsq} layer can be added after triplane creation to learn a discrete codebook and produce discrete tokens.
However, in initial experiments, we found that discrete tokens underperformed continuous tokens.
Thus, only continuous tokens are used in this work.

\section{Experiments}
\label{sec:expts}

\begin{figure*}[t]
    \centering
    \includegraphics[width=\linewidth]{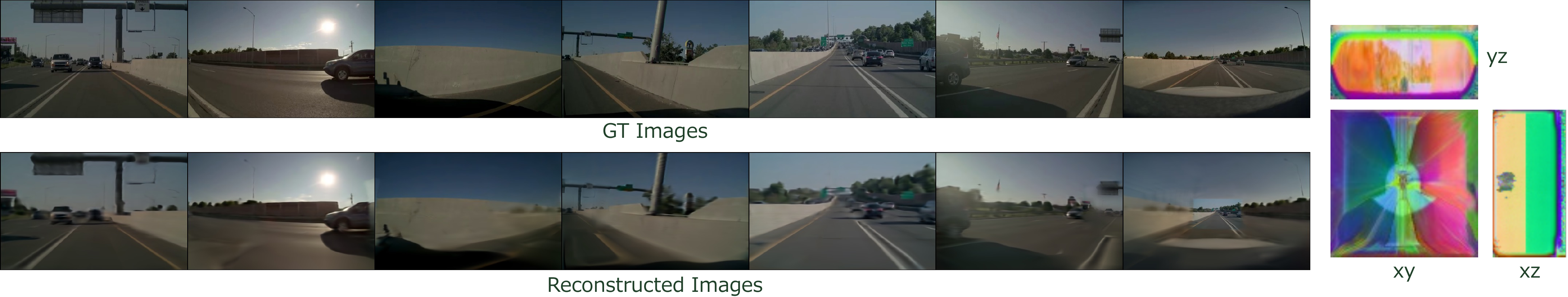}
    
    \vspace{-0.2cm}
    
    \caption{Triplanes can accurately represent environments observed by outward-facing cameras (\textbf{top}), producing accurate reconstructions (\textbf{below}) and semantically-meaningful features after only being trained with self-supervised pixel reconstruction losses (\textbf{right}, a PCA visualization of triplane features).}
    \label{fig:triplane_results}
    \vspace{-0.4cm}
\end{figure*}

\textbf{Dataset.} To train and evaluate driving performance, we leverage a large internal dataset
consisting of 20,000 hours of driving data from multiple ego-vehicles in 1700+ cities and 25 countries. Accordingly, it contains a variety of driving scenarios including highway and urban driving, multiple weather conditions, day and night times, and varying amounts of traffic, with a geographically-separate 90\% trainval and 10\% test split. Each ego-vehicle has 7 cameras which we downsample to a per-camera resolution of $H \times W = 320 \times 512$ with 10 Hz observation frequency. \cref{fig:triplane_results} visualizes an example from the dataset.

\begin{figure}[t]
    \centering
    \includegraphics[width=\linewidth]{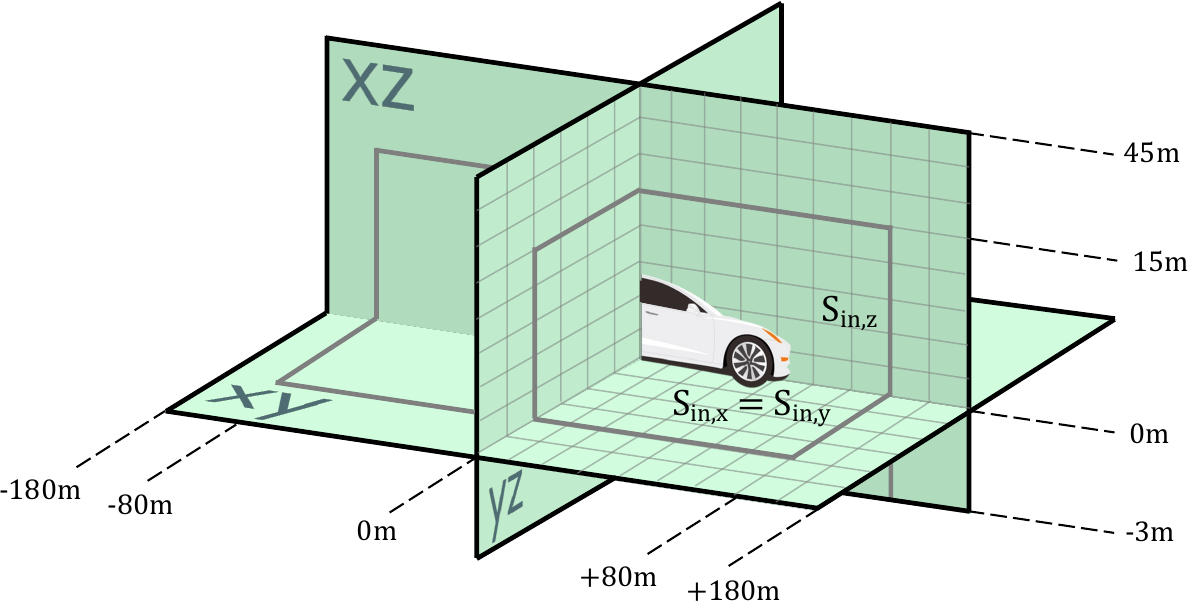}
    \caption{We employ a bilinear triplane resolution, representing regions outside of $S_{\text{in},x}, S_{\text{in},y}, S_{\text{in},z}$ with fewer grid cells.}
    \label{fig:triplane_sizes}

    \vspace{-0.4cm}
\end{figure}

\textbf{Triplane Model.} In all experiments, we set $S_x = S_y = 96$, $S_z = 48$, and $D_f = 192$. To account for driving in both high-speed freeways and slower urban streets, each triplane metrically represents 180m ahead, behind, left, and right of the ego-vehicle, as well as 45m above and 3m below the ego-vehicle, visualized in \cref{fig:triplane_sizes}.
As described in \cref{sec:method_encoding}, we employ a symmetric bilinear grid resolution with $S_{\text{in},x} = S_{\text{in},y} = 36$ cells and, since we cannot see under ground, the $z$-axis asymmetrically represents $[-3, 15]$m with 36 cells and $(15, 45]$m with the remaining 12 cells.

We use the 22M-parameter DINOv2-small~\cite{oquab2024dinov} model for ImageEnc in \cref{eqn:featurize} and two sets of per-image and cross-image attention layers (4 total attention layers, totaling 6.5M parameters) to convert image features to triplanes.

\textbf{AR Transformer.} An LLM-like backbone serves as our representative AR Transformer model~\cite{wu2025alpamayo}. It takes as input the above sensor tokens as well as past ego trajectory information to produce future ego-vehicle positions. Future trajectories are represented as discrete tokens and decoded to positions at 10 Hz.

\textbf{Baselines.} Since image autoencoders and ViTs are the most commonly-used image tokenizers (\cref{sec:related_work}), we compare our work to VQGAN~\cite{esser2021vqgan} and DINOv2-small~\cite{oquab2024dinov} as representative models. 

\textbf{Training.} We employ a multi-stage training pipeline. Tokenizers are first pre-trained to model driving data on our internal dataset (except for DINOv2 as it is already internet-pretrained), followed by training the randomly-initialized backbone on the resulting tokens. The overall tokenizer-LLM combination is trained to minimize a next-trajectory-token prediction loss. We set $\lambda_\text{LPIPS} = \lambda_1 = 0.5$ in \cref{eqn:loss}. Additional training details can be found in~\cite{wu2025alpamayo}.

\textbf{Metrics.} To evaluate triplane quality, we compare the reconstructed images to GT images using peak signal-to-noise ratio (PSNR) and structural similarity index measure (SSIM).
To evaluate the combined tokenizer-LLM open-loop driving performance, we leverage the commonly-used minADE$_6$ metric (denoting the minimum average displacement error among 6 sampled trajectories), formulated as
\begin{equation}
    \text{minADE}_k(\hat{o}, o) = \min_{k=1,\dots,6} \frac{1}{T} \sum_{t=1}^T \| \hat{o}_k^{(t)} - o^{(t)}\|_2
\end{equation}

\subsection{Triplane Representation Quality}

\begin{table}
  \centering
  \begin{tabular}{l|cc}
    \toprule
    Choice of ImageEnc in \cref{eqn:featurize} & PSNR $\uparrow$ & SSIM $\uparrow$ \\
    \midrule
    DINOv2-small~\cite{oquab2024dinov} (4 cameras) & 29.15 & 0.85 \\
    ResNet50~\cite{HeZhangEtAl2016} (4 cameras) & 29.58 & 0.86 \\
    \midrule
    DINOv2-small~\cite{oquab2024dinov} (7 cameras) & 28.94 & 0.84 \\
    ResNet50~\cite{HeZhangEtAl2016} (7 cameras) & 29.34 & 0.85 \\
    \bottomrule
  \end{tabular}
  \caption{Irrespective of the choice of image encoder or number of cameras, triplanes can accurately model outdoor driving scenes, matching state-of-the-art neural reconstruction performance.}
  \label{tab:triplane_rep_perf}
  \vspace{-0.4cm}
\end{table}

To evaluate whether triplanes can represent driving scenarios collected from outward-facing cameras, we first evaluate their reonstruction performance after the first stage of training. \cref{fig:triplane_results} visualizes triplane-rendered images, GT images, and a 3-color triplane feature visualization (obtained with PCA). As can be seen in \cref{tab:triplane_rep_perf}, triplanes are able to represent both nearby and far-away elements of the scene, achieving 28.94 dB PSNR and 0.84 SSIM when modeling all 7 cameras, comparable to state-of-the-art neural rendering performance~\cite{yang2025storm}. If modeling fewer cameras (e.g., only the front 4 cameras), reconstruction performance increases to 29.15 dB PSNR and 0.85 SSIM, since more modeling capacity can be used per-image.

\textbf{Alternative Choices for ImageEnc.} Our work can operate with any image backbone that produces 2D features. To demonstrate this, \cref{tab:triplane_rep_perf} also shows triplane reconstruction performance with a CNN-based image encoder (an ImageNet-pretrained ResNet50~\cite{HeZhangEtAl2016}). As can be seen, this choice yields even better PSNR and SSIM values, although differences in model optimization result in an increase in inference runtime compared to DINOv2-small. Accordingly, we use DINOv2-small for ImageEnc in \cref{eqn:featurize} in all subsequent experiments.

\subsection{Open-Loop Driving Performance}

\begin{table}
  \centering
  \begin{tabular}{lc|ccc}
    \toprule
    & Tokens per & \multicolumn{3}{c}{minADE$_6$ (m) $\downarrow$}\\
    Image Tokenizer & Image $\downarrow$ & @1s & @3s & @5s \\
    \midrule
    VQGANEnc & 160 & \textbf{0.08} & 0.33 & 0.74 \\
    DINOv2-small & 160 & \textbf{0.08} & 0.32 & 0.69 \\
    \midrule
    Ours (8-8-8) & \textbf{45} & \textbf{0.08} & 0.32 & 0.72 \\
    Ours (4-6-6) & 104 & \textbf{0.08} & \textbf{0.31} & \textbf{0.67} \\
    \bottomrule
  \end{tabular}
  \vspace{-0.2cm}
  \caption{When paired with a 1B AR backbone, our triplane-based tokenizer achieves similar open-loop motion planning performance as baseline tokenizers, while producing significantly fewer sensor tokens. Patch sizes are denoted as $(p_x-p_y-p_z)$.}
  \label{tab:llama_perf}
  \vspace{-0.4cm}
\end{table}

\begin{table}[th]
  \centering
  \setlength{\tabcolsep}{4.25pt}
  \small
  \begin{tabular}{l|cccc}
    \toprule
    \textbf{nuScenes E2E Planning} & \multicolumn{4}{c}{Traj L2 (m) $\downarrow$}\\
     & @1s & @2s & @3s & Ave$_{1,2,3s}$ \\
    \midrule
    UniAD~\cite{hu2023planning} & 0.48 & 0.89 & 1.47 & 0.95\\
    VAD-Base~\cite{jiang2023vad} & 0.41 & 0.86 & 1.46 & 0.91\\
    PARA-Drive~\cite{weng2024paradrive} & 0.26 & 0.59 & 1.12 & 0.66\\
    TOKEN 7B~\cite{tian2024token} & 0.26 & 0.70 & 1.46 & 0.81\\
    DiMA 7B (VAD-Tiny)~\cite{hegde2025dima} & 0.20 & 0.53 & 1.10 & 0.61\\
    DiMA 7B (VAD-Base)~\cite{hegde2025dima} & 0.18 & 0.50 & 1.02 & 0.57\\
    \midrule
    DINOv2-small + 1B & 0.34 & 0.69 & 1.17 & 0.73\\
    Ours (8-8-8) + 1B & 0.31 & 0.63 & 1.10 & 0.68\\
    Ours (4-6-6) + 1B & 0.29 & 0.62 & 1.08 & 0.66\\
    \bottomrule
  \end{tabular}
  \vspace{-0.2cm}
  \caption{Our approach is competitive with state-of-the-art approaches, even with a much smaller backbone (1B) and much fewer input tokens ($35\%$ to $72\%$ less). Triplane patchification sizes are denoted with $(p_x-p_y-p_z)$.}
  \label{tab:nuscenes}
  \vspace{-0.4cm}
\end{table}

To evaluate driving performance with model sizes that are suitable for deployment, we train a 1B-parameter LLM-like model ($D_{\text{AR}} = 2048$) to drive from the 4 front-facing cameras, with their past 6 frames as scene context (totaling 24 frames of input). In addition to the VQGAN and DINOv2-small tokenizer baselines, we evaluate two patch size $p_x, p_y, p_z$ ablations for our triplane tokenizer. The first, $(p_x, p_y, p_z) = (4, 6, 6)$, yields 416 tokens per timestep (104 tokens per image, 35\% fewer than the baselines) and matches the inference runtime of the baseline approaches. The second employs more aggressive triplane patchification, with $(p_x, p_y, p_z) = (8, 8, 8)$, resulting in 180 tokens per timestep (45 tokens per image, 72\% fewer than the baselines) and a significantly faster runtime. As can be seen in \cref{tab:llama_perf}, the backbone trained with our approach performs similarly to the baselines across all timescales while requiring significantly fewer tokens per image.

In~\cref{tab:nuscenes}, we provide a comparison of our approach and the DINOv2 baseline to a variety of other state-of-the-art (SotA) approaches for E2E planning on the nuScenes~\cite{CaesarBankitiEtAl2019} validation set. We follow the standardized evaluation setup from PARA-Drive~\cite{weng2024paradrive}: predicting 3s of future ego-motion from 2s of past image frames, training only on the nuScenes train set, and using the same displacement metric definitions.
As can be seen, our work is competitive with SotA approaches. Further, our approach performs similarly to works like DiMA~\cite{hegde2025dima} and TOKEN~\cite{tian2024token} (which use much larger 7B LLM backbones compared to our 1B), while running much faster due to our reduced input token counts. Importantly, both the (4-6-6) and (8-8-8) versions of our approach outperform the baseline DINOv2-based tokenizer.
Further open-loop evaluations can be found in the appendix.

\begin{figure*}[t]
    \centering
    \begin{subfigure}{\linewidth}
        \centering
        \includegraphics[width=\linewidth]{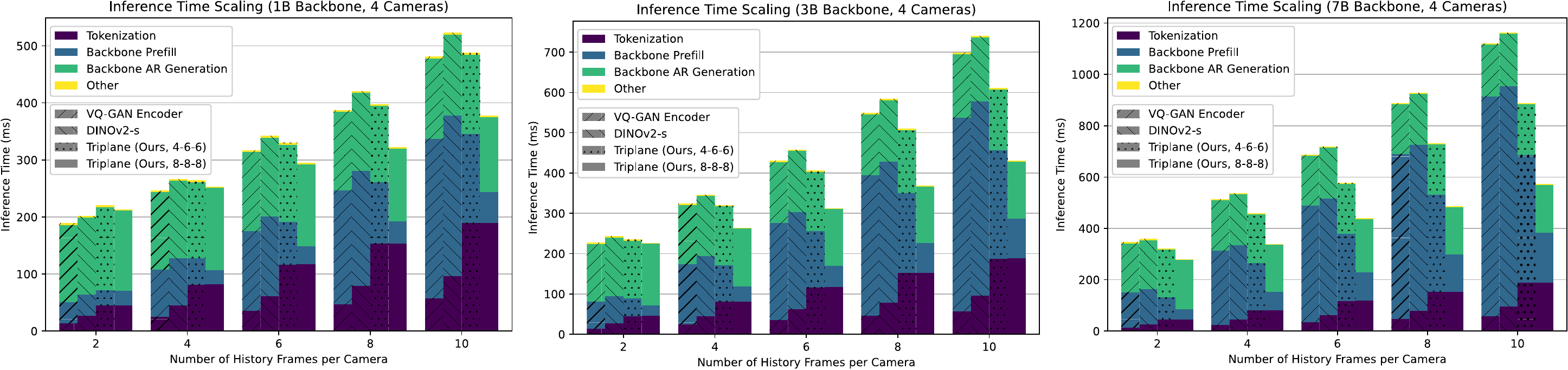}
        \caption{Inference time scaling results when taking the 4 front-facing cameras as input.}
    \end{subfigure}
    
    \vspace{0.175cm}
    
    \begin{subfigure}{\linewidth}
        \centering
        \includegraphics[width=\linewidth]{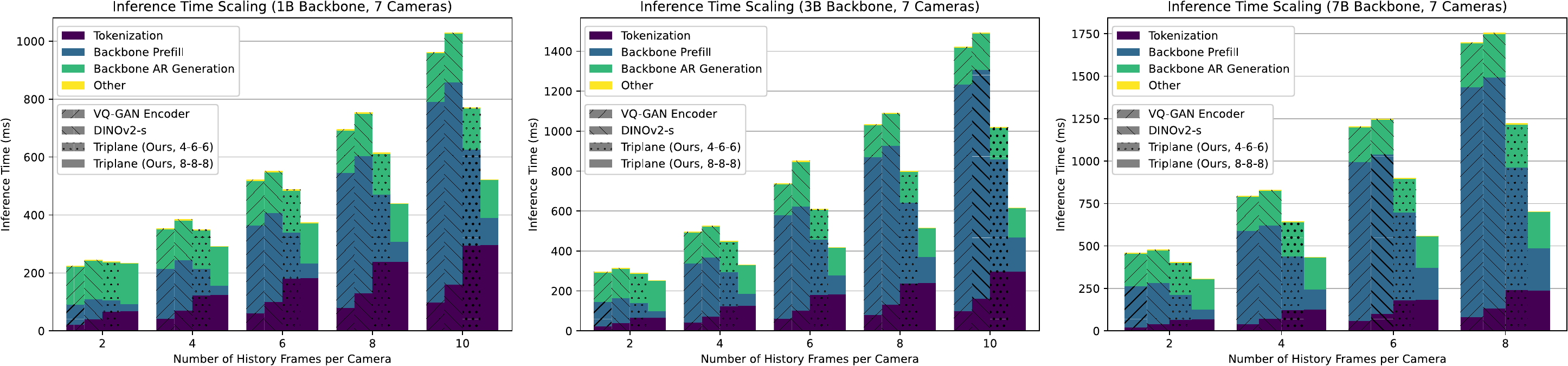}
        \caption{Inference time scaling results when taking all 7 cameras as input.}
    \end{subfigure}

    \vspace{-0.2cm}
    
    \caption{Our triplane-based multi-camera tokenization approach scales much more favorably than baseline approaches as the number of cameras and number of context frames per camera increases. This trend holds across multiple autoregressive Transformer sizes.}
    \label{fig:runtime_scaling}
    \vspace{-0.4cm}
\end{figure*}

\begin{figure}[t]
    \centering
    \includegraphics[width=0.5\linewidth]{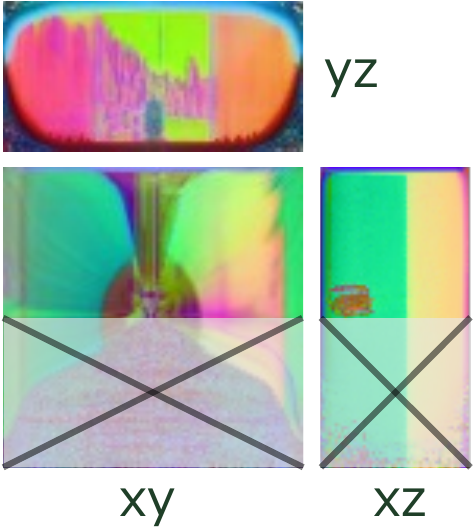}

    \vspace{-0.3cm}
    
    \caption{If modeling cameras that face a certain direction (e.g., all front-facing cameras, as in this figure), the number of tokens can be further reduced by removing unused triplane regions, indicated with an ``x". Depending on the chosen patch sizes $p_x, p_y, p_z$, this can further reduce token counts by up to 40\%.}
    \label{fig:halfplane}
    \vspace{-0.4cm}
\end{figure}

\textbf{Halfplane Token Reduction.} Note that, in this particular setup (modeling only the front-facing cameras), one can further reduce the number of tokens produced by our triplane-based approach by removing half of the $xy$ and $xz$ planes (they model space behind the car, and are thus unused, as visualized in \cref{fig:halfplane}). The results in \cref{tab:llama_perf} use this token-reduction technique, as do all subsequent experiments modeling only the 4 front-facing cameras. Depending on the chosen patch sizes, this can further reduce token counts by 30-40\% (specifically, 37.5\% for $(p_x, p_y, p_z) = (4, 6, 6)$ and 40.9\% for $(p_x, p_y, p_z) = (8, 8, 8)$).

\subsection{Inference Time Scaling}

Our main hypothesis for using a triplane-based multi-camera tokenization strategy is that, while adding any layers to create an intermediate representation will increase tokenization runtime, the corresponding decrease in sensor tokens will lead to greater savings in backbone runtime, resulting in an overall inference time decrease since billion-parameter Transformers are virtually always the main inference bottleneck in such AV stacks.

To confirm this hypothesis, we profile the runtimes of combined tokenizer-backbone models with a variety of backbone model sizes, number of historical frames, and numbers of cameras. In addition to the 1B-parameter version above, we also profile the runtime of 3B-parameter and 7B-parameter backbones. Such model sizes were previously considered for AV deployment, and even successfully demonstrated on real AVs in works such as~\cite{tian2024drivevlm}. We report the average inference times of different parts of the combined model over 100 forward passes on a single A100 80GB GPU in \cref{fig:runtime_scaling}. We omit error bars as there is very little variation across runs (95\% confidence intervals are within 0.5ms of means).

\begin{figure*}[t]
  \centering
   \includegraphics[width=\linewidth]{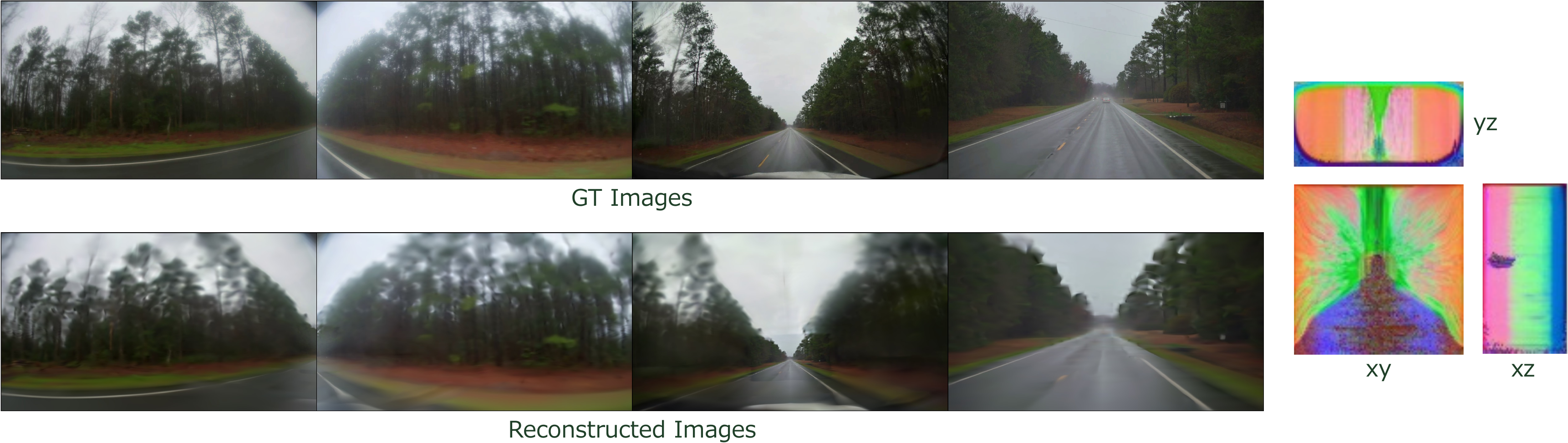}
   \caption{Triplanes can faithfully represent higher resolution data without requiring any changes to triplane sizes. Since these cameras are all front-facing, the rear half of the $xy$ and $xz$ planes are unused.}
   \label{fig:highres}
   \vspace{-0.2cm}
\end{figure*}

As can be seen in \cref{fig:runtime_scaling}, the majority of overall inference time is spent in the backbone (blue and green bars) and we observe the expected linear scaling of inference time as the number of cameras increases (from the top row to the bottom row) and as the number of context frames per camera increases (within each plot).

\cref{fig:runtime_scaling} confirms that our method demonstrates the tradeoff described above: an increase in tokenization time (larger purple bar) yields a much faster AR prefill stage (much smaller blue bar), with additional gains if using a more aggressive patchification scheme (increasing $p_x, p_y, p_z$). This results in our method achieving significant inference time reductions compared to baselines, often \emph{halving} model runtime with larger context sizes and number of cameras.

If targeting a 3 Hz inference frequency (e.g., as in~\cite{tian2024drivevlm}), it is difficult to incorporate more than 2-4 frames of context with model sizes above 1B parameters when using baseline tokenization approaches. In contrast, our triplane-based tokenizer enables running a 7B-parameter model with 7 cameras and 4 frames of context at 3 Hz!

An additional benefit of triplane-based tokenization is that patchification can be modified after training, since it is executed \emph{after} triplane creation, enabling easier tuning of overall model performance without retraining the triplane.

\subsection{Resolution and Camera Count Agnosticism}

\textbf{Higher Resolutions.} To validate that our approach can seamlessly handle higher image resolutions, we resample images from our dataset with a $3.6 \times$ higher resolution, $H' \times W' = 576 \times 1024$, and retrain our triplane-based approach with them. The resulting model achieves 28.00 dB PSNR and 0.82 SSIM, which is lower than the lower-resolution results in \cref{tab:triplane_rep_perf}. This is to be expected, however, as we do not alter the triplane size, meaning $3.6\times$ more pixels are being modeled by the same number of features. \cref{fig:highres} confirms qualitatively that triplanes can faithfully represent higher resolution data, without requiring any changes to triplane sizes.

To fully verify that the resulting tokens are still useful for driving, we retrain the 1B backbone on the resulting tokens (using $(p_x, p_y, p_z) = (4, 6, 6)$) and observe identical performance to our previous $(4, 6, 6)$ model in \cref{tab:llama_perf}.

\begin{figure}[t]
  \centering
   \includegraphics[width=\linewidth]{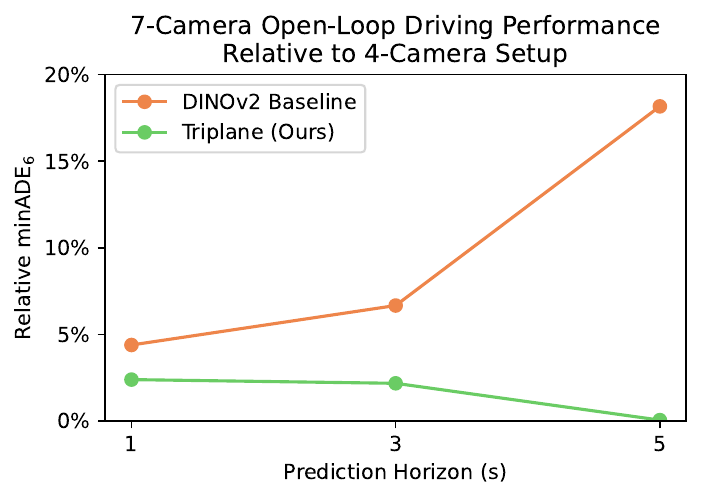}

   \vspace{-0.2cm}

   \caption{Our triplane model performs similarly with 7 cameras and 4 cameras, whereas the DINOv2-small baseline struggles.
   }
   \label{fig:llama_7cam}
   \vspace{-0.4cm}
\end{figure}

\textbf{Camera Count Agnosticism.} To confirm our method's agnosticism to the number of cameras, as initially demonstrated in \cref{tab:triplane_rep_perf}, we measure the motion planning performance of the 1B backbone with 7 cameras and compare it to the 4-camera equivalent from above. \cref{fig:llama_7cam} shows an interesting trend: When paired with our 7-camera triplane-based approach, the 1B backbone is able to achieve performance within $2.5\%$ of its 4-camera performance. However, when paired with the DINOv2-small baseline, it yields $18\%$ worse open-loop motion planning performance. We also observed training instabilities with the 7-camera DINOv2-small baseline, potentially due to the long sequence length brought by modeling 7 cameras each with 6 context frames (totaling 42 frames of input).

\begin{figure*}[t]
  \centering
  \includegraphics[width=0.495\linewidth]{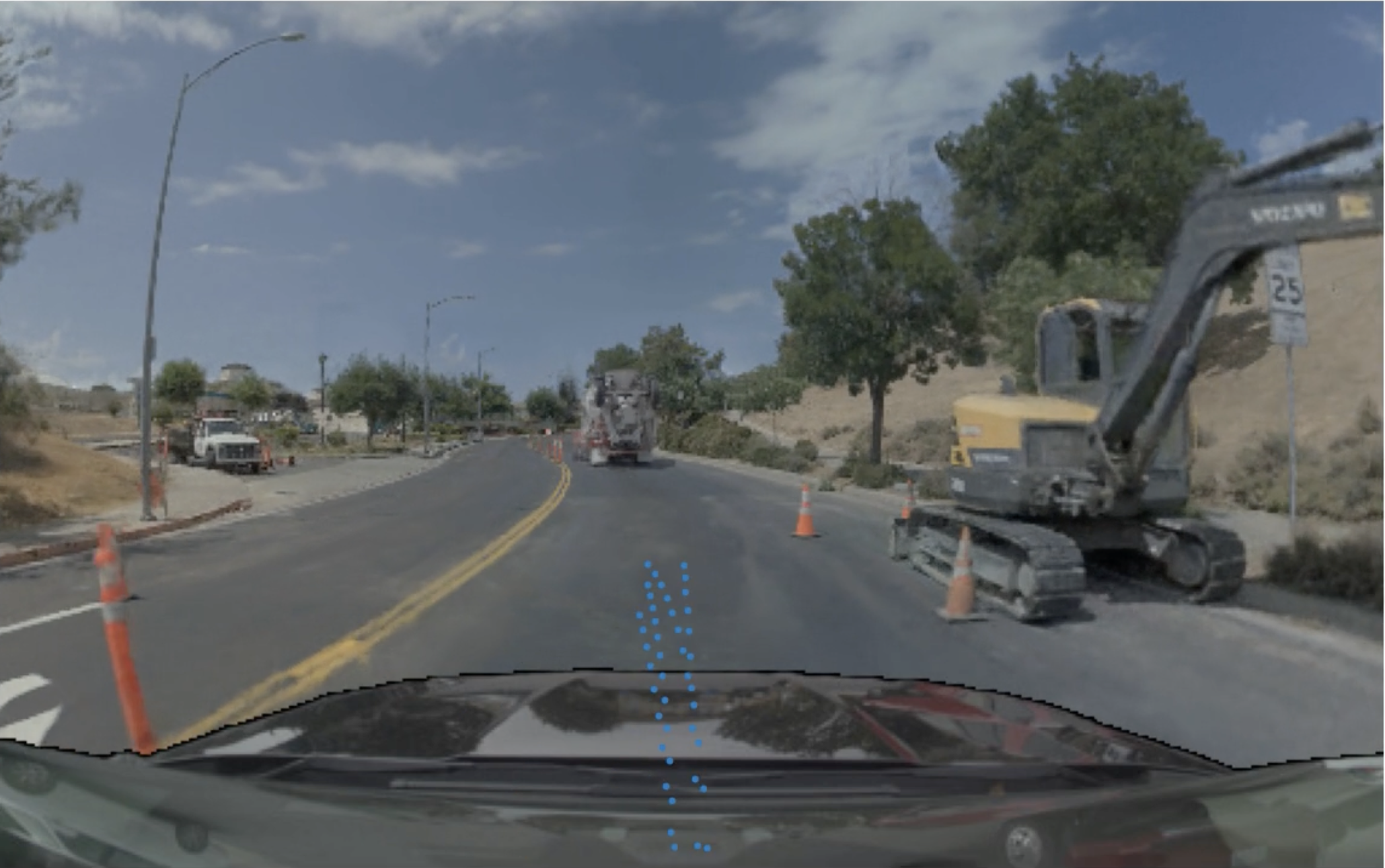}
  \hfill
  \includegraphics[width=0.495\linewidth]{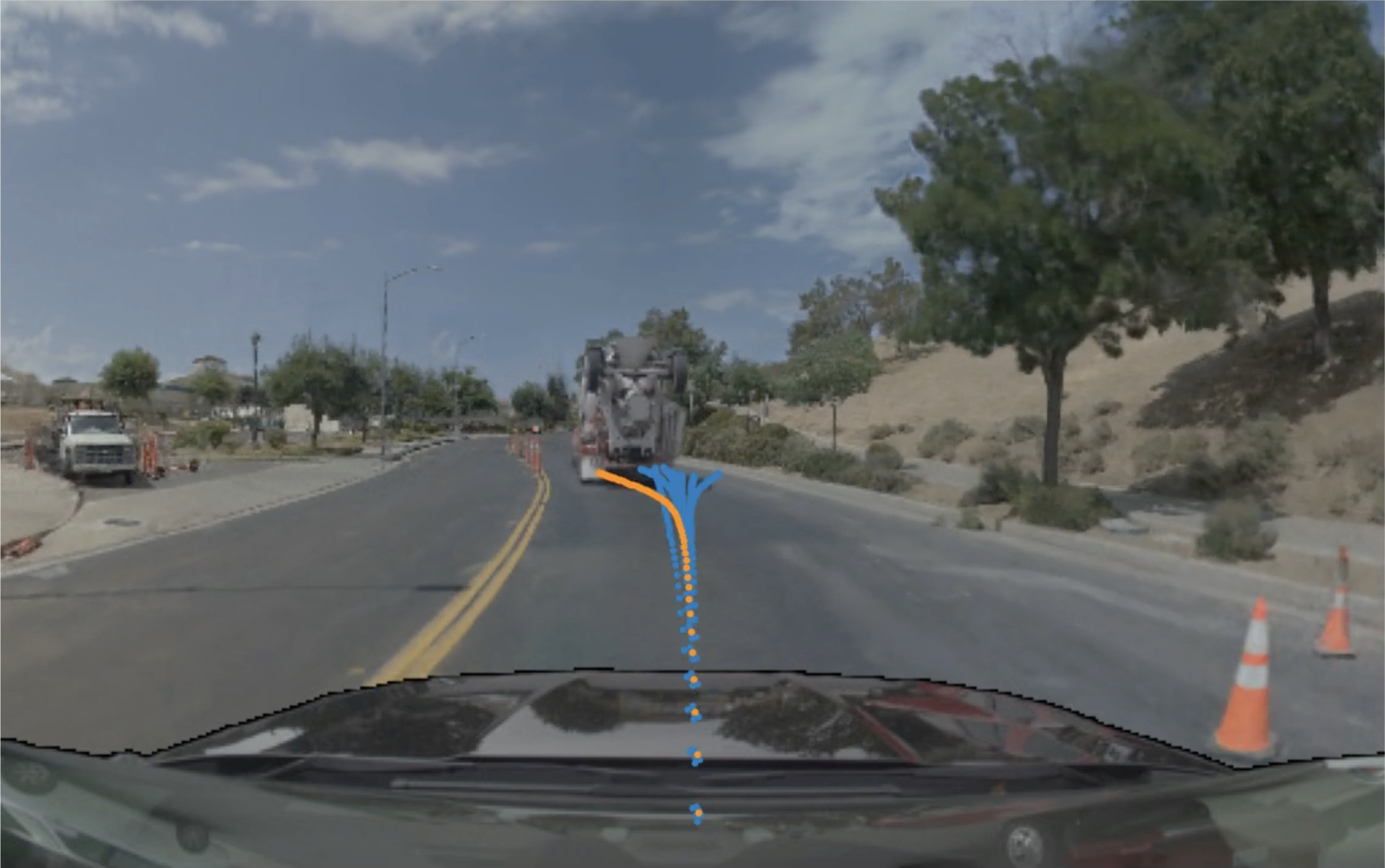}
   \caption{In a challenging closed-loop simulation of a construction scene, the baseline DINOv2-based model (\textbf{left}) stops in the middle of the road and does not start driving again, whereas our triplane-based model (\textbf{right}) has no difficulty nudging around the parked excavator and driving behind the cement mixer. On the front camera views are the predicted ego-vehicle trajectories (selected driving trajectory in {\color{orange} orange}, with the other 5 predicted trajectories in {\color{blue} blue}).}
   \label{fig:cl_sim}
   \vspace{-0.2cm}
\end{figure*}

\subsection{Performance in Closed-Loop Simulation}

Finally, it is well-known that open-loop driving performance does not necessarily correlate to closed-loop driving performance~\cite{Dauner2024NEURIPS}. Accordingly, we additionally validate the driving performance of the combined tokenizer-LLM models in a closed-loop simulation environment.

We conduct closed-loop evaluation using a state-of-the-art in-house neural reconstruction (NR)-based simulator, capable of reconstructing logged driving events and rendering novel views if the ego-vehicle departs from its originally-logged trajectory~\cite{wu2025alpamayo}. We place the combined tokenizer-LLM models within 75 challenging scenarios, mined specifically due to the amount of ego-agent and agent-agent interactions within them, each of which are 20s-long. \cref{fig:cl_sim} shows one such rollout with the currently-predicted trajectories superimposed over the image. Within these scenarios, we have access to high-quality GT map labels, and so we compare models by their offroad rates (in what proportion of closed-loop rollouts did the ego-vehicle leave the road). Owing to the significant speedups achieved by the $(p_x, p_y, p_z) = (8, 8, 8)$ version of our triplane-LLM combined model, we evaluate it against the DINOv2-LLM combined baseline. \cref{tab:cl_sim} shows that our approach achieves a slightly lower offroad rate compared to the baseline.

\begin{table}
  \centering
  \begin{tabular}{lc|ccc}
    \toprule
    & Tokens per & \multicolumn{3}{c}{Offroad Rate $\downarrow$}\\
    Image Tokenizer & Image $\downarrow$ & @6s & @12s & @20s \\
    \midrule
    DINOv2-small & 160 & 2.7\% & 2.7\% & 4.0\% \\
    Ours (8-8-8) & 45 & \textbf{1.4\%} & \textbf{1.4\%} & \textbf{2.7\%} \\
    \bottomrule
  \end{tabular}
  \caption{When paired with a 1B AR backbone, our triplane-based tokenizer achieves slightly better closed-loop driving performance as the DINOv2-small baseline, while producing significantly fewer sensor tokens. Patch sizes are denoted $(p_x-p_y-p_z)$.}
  \label{tab:cl_sim}
  \vspace{-0.4cm}
\end{table}

\section{Limitations and Conclusion}
\label{sec:conclusion}

In conclusion, we present a novel multi-camera image tokenization scheme based on triplanes that can produce sensor tokens in a resolution-agnostic, camera-number-agnostic, and geometrically-aware manner. In comparison to autoencoder and ViT-based baselines, our triplane-based approach is able to represent multi-camera images with much fewer tokens, yielding significant inference time reductions while achieving comparable open-loop motion planning performance and even slightly improved offroad rates in closed-loop simulations.

A core limitation of the proposed approach is that it focuses on per-timestep triplane modeling, leaving across-timestep modeling strategies to future work. Other exciting areas of future work include further triplane size reduction and feature dimension compression (to limit memory requirements), and further reductions in tokenizer latency (potentially leveraging model optimization tools such as pruning and optimized runtime environments, e.g., TensorRT), with the ultimate goal of deploying triplane-based multi-camera tokenization on-vehicle.

\paragraph{Acknowledgments.}
We thank Jiawei Yang, Yue Wang, Seung Wook Kim, S\'{e}rgio Agostinho, Amrita Mazumdar, Omer Shapira, and Shalini De Mello for many helpful technical discussions and guidance, as well as Michael Watson, Maximilian Igl, Michal Tyszkiewicz, Aaron Smith, Peter Karkus, and Zan Gojcic for their support in executing closed-loop simulations.

{
    \small
    \bibliographystyle{ieeenat_fullname}
    \bibliography{main}
}

\clearpage
\setcounter{page}{1}
\appendix
\maketitlesupplementary

\section{Additional Open-Loop Evaluations}
\label{sec:supp_waymo}

\begin{table}[th]
  \centering
  \small
  \begin{tabular}{l|cccc}
    \toprule
    \textbf{WOD E2E Planning} & \multicolumn{4}{c}{Traj L2 (m) $\downarrow$}\\
     & @1s & @3s & @5s & Ave$_{1,3,5s}$ \\
    \midrule
    DINOv2-small + 1B & 0.21 & 1.00 & 2.31 & 1.17\\
    Ours (8-8-8) + 1B & \textbf{0.11} & \textbf{0.66} & \textbf{1.72} & \textbf{0.83}\\
    Ours (4-6-6) + 1B & \textbf{0.11} & 0.69 & 1.79 & 0.86\\
    \bottomrule
  \end{tabular}
  \vspace{-0.3cm}
  \caption{Our approach significantly outperforms the DINOv2 baseline on the Waymo Open Dataset. Triplane patchification sizes are denoted with $(p_x-p_y-p_z)$.}
  \label{tab:waymo}
\end{table}

\cref{tab:waymo} shows the performance of our work on the Waymo Open Dataset~\cite{SunKretzschmarEtAl2020} validation split, predicting 5s of future ego-motion from 1s of past observations (with the same metrics as above at 1s, 3s, 5s horizons). Here, the DINOv2-based tokenizer performs much worse than our work, indicating that the non-geometry-aware ViT-based tokenization struggles to capture the details needed to produce the more complicated driving behaviors found in Waymo.

Interestingly, we also see that our planning performance is preserved even with aggressive patchification schemes ((8-8-8) slightly outperforms (4-6-6)!), indicating that our triplane-based tokenizer features are robust to downstream patchification choices. Unfortunately, there do not appear to be many other works that utilize Waymo for E2E planning benchmarking, so we offer these results primarily as further evidence that our conclusions generalize to other datasets.

\section{Optional Auxiliary Supervision}
\label{sec:supp_auxsup}

\begin{figure}[ht]
  \centering
  \includegraphics[width=\linewidth]{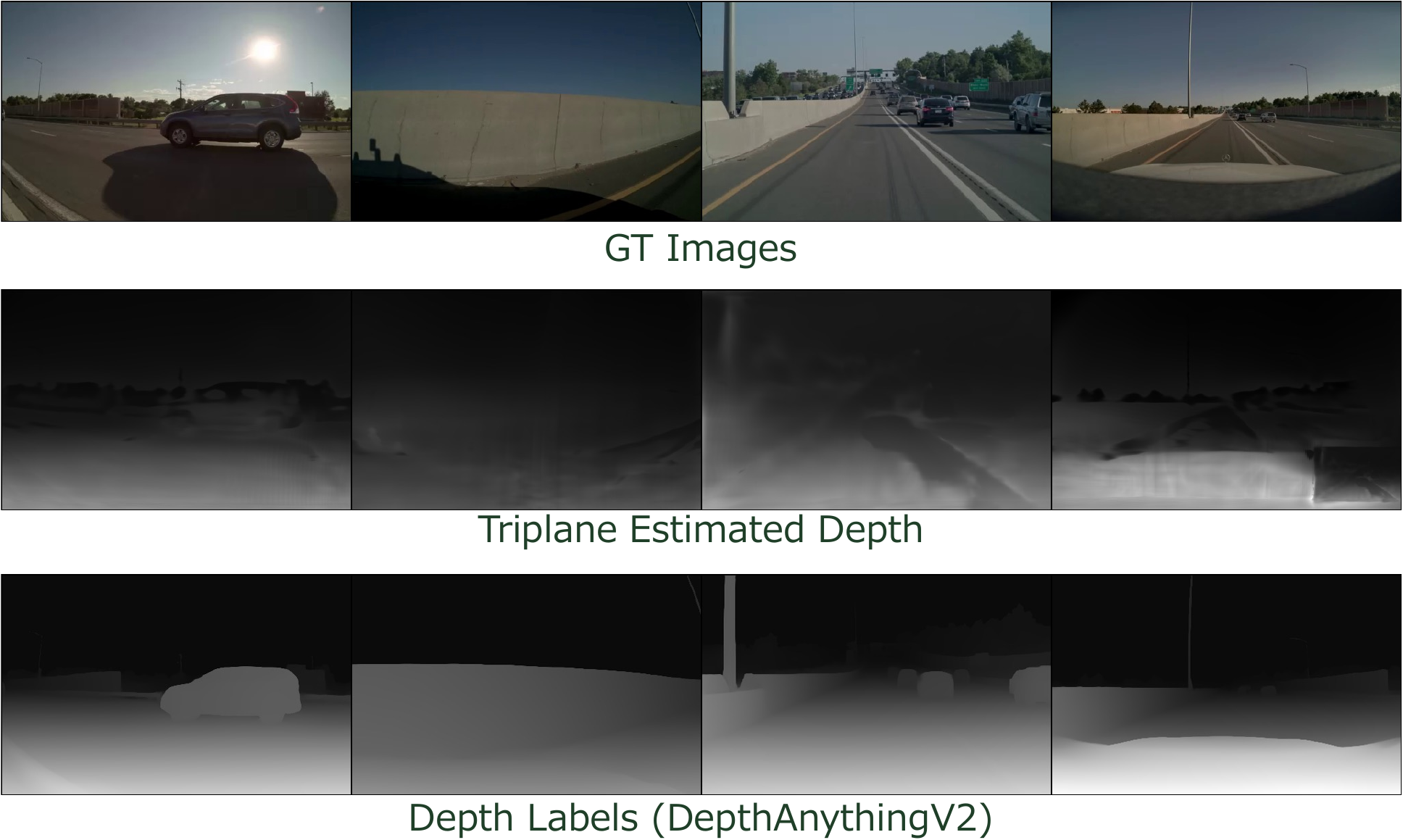}
   \caption{Triplanes can optionally be trained with auxiliary labels, if available. Here, DepthAnythingV2~\cite{depth_anything_v2} provides relative depth supervision which the triplane is able to learn from and predict. As there are many high-frequency details in these images, they are best viewed on a high-resolution screen.}
   \label{fig:depth}
\end{figure}

If additional sources of labeled data are available, e.g., as part of prior AV development efforts, they can easily be incorporated into training as supplementary terms to the self-supervised rendering loss (\cref{eqn:loss}). In \cref{fig:depth}, we show the results of adding a depth supervision loss to \cref{eqn:loss}. Specifically, we use DepthAnythingV2-Small~\cite{depth_anything_v2} to provide relative depth labels, and include them in the loss through an additional depth term $\mathcal{L}_{\text{depth}}(d, \hat{d}) = \lambda_d \|d - \hat{d}\|_1$ where $d$ is the provided depth label and $\hat{d}$ is the triplane-estimated depth. On the left-most image of \cref{fig:depth}, a car can be seen two lanes over from the ego-vehicle. This car is also visible in the triplane-estimated depth, with more details (e.g., the wheels and the shape of the driver-side windows are visible). While these preliminary results are encouraging, and validate the ability for multi-task training with triplanes, we leave more accurate depth (and other auxiliary tasks) modeling to future work.

\end{document}